%% file: acl2023.tex
\pdfoutput=1
\documentclass[11pt]{article}

\usepackage{ACL2023}
\usepackage{times}
\usepackage{latexsym}

\usepackage[T1]{fontenc}
\usepackage[utf8]{inputenc}

\usepackage{microtype}

\usepackage{inconsolata}

\usepackage{bm, amsfonts, amsmath}
\usepackage{hyperref}
\usepackage{booktabs}
\usepackage{multirow}
\usepackage{graphicx}
\usepackage{tabularx}
\usepackage{array}
\usepackage{microtype}
\usepackage{algorithm, algpseudocode}
\newcommand{\metric}{FIZZ}
\newcommand{\ie}{\textit{i}.\textit{e}.}
\renewcommand*{\thefootnote}{\fnsymbol{footnote}}
\title{FIZZ: Factual Inconsistency Detection by Zoom-in Summary and Zoom-out Document}
\author{
Joonho Yang$^{1}$,~Seunghyun Yoon$^{2}$,~Byeongjeong Kim$^{1}$,~Hwanhee Lee$^{1}$\textsuperscript{$\dagger$} \\
    $^{1}$Department of Artificial Intelligence, Chung-Ang University, $^{2}$Adobe Research, USA\\
    \texttt{\{plm3332, michael97k, hwanheelee\}@cau.ac.kr}, \texttt{syoon@adobe.com}
}

\begin{document}
\maketitle
% Anonymous Authors
\footnotetext{\textsuperscript{$\dagger$}Corresponding author.}
\renewcommand*{\thefootnote}{\arabic{footnote}}
\input{0_abstract}
\input{1_intro}
\input{2_relatedwork}
\input{3_method}
\input{4_experiments}
\input{5_conclusion}
\input{6_limitations}
\input{7_ethics}
\input{8_acknowledgement}

\bibliography{main}
\bibliographystyle{acl_natbib}
\input{9_appendix}
\end{document}

%% file: 0_abstract.tex
\begin{abstract}
%Recent developments in the field of text summarization systems, particularly through the advent of Large Language Models (LLMs), have demonstrated remarkable advancements in performance.
Through the advent of pre-trained language models, there have been notable advancements in abstractive summarization systems.
% However, a notable challenge persists as a substantial number of automatically generated summaries exhibit factual inconsistencies, such as hallucinations.
%Concurrently, the tendency of these summarization systems to introduce factual inaccuracies has been the subject of extensive research, including the development and application of metrics aimed at detecting factual inconsistency errors.
Simultaneously, a considerable number of novel methods for evaluating factual consistency in abstractive summarization systems has been developed. But these evaluation approaches incorporate substantial limitations, especially on refinement and interpretability.
In this work, we propose highly effective and interpretable factual inconsistency detection method \textbf{\metric} (\textbf{F}actual \textbf{I}nconsistency Detection by \textbf{Z}oom-in Summary and \textbf{Z}oom-out Document) for abstractive summarization systems that is based on fine-grained \textit{atomic facts} decomposition.
Moreover, we align \textit{atomic facts} decomposed from the summary with the source document through adaptive granularity expansion.
These \textit{atomic facts} represent a more fine-grained unit of information, facilitating detailed understanding and interpretability of the summary's factual inconsistency.
Experimental results demonstrate that our proposed factual consistency checking system significantly outperforms existing systems.
% We release the code at \url{https://anonymous.4open.science/r/FAIRY-B5F7}.
We release the code at \url{https://github.com/plm3332/FIZZ}.
\end{abstract}

%% file: 1_intro.tex
\section{Introduction}
% 연구 주제 및 지금까지의 연구의 문제점
%In the era of Large Language Models (LLMs), the development of automatic summarization systems leveraging LLMs has achieved remarkable advancements in producing fluent and natural summaries~\cite{chang2023survey}. 
With the development of pre-trained language models, abstractive summarization systems using these language models have made remarkable progress in generating fluent and natural summarizations~\cite{chang2023survey}.
However, one of the notable challenges these systems confront is the hallucination, causing language models to generate summaries that are factually inconsistent with the given article~\cite{maynez-etal-2020-faithfulness, kryscinski-etal-2020-evaluating, tam-etal-2023-evaluating, Zhang2023SirensSI}. 
Recognizing the significance of this issue, various evaluation metrics have been introduced to detect these errors, starting from traditional methods like ROUGE~\cite{lin-2004-rouge} and BERTScore~\cite{zhang2020bertscore} to a large number of advanced metrics~\cite{goyal-durrett-2020-evaluating, goyal-durrett-2021-annotating, scialom-etal-2021-questeval, fabbri-etal-2022-qafacteval, laban-etal-2022-summac, luo2023chatgpt, zha-etal-2023-alignscore, wang-etal-2023-chatgpt}. Especially, many of the recent works~\cite{laban-etal-2022-summac, schuster-etal-2022-stretching, zha-etal-2023-alignscore} adopted sentence level evaluation using Natural Language Inference (NLI) systems for factual consistency checking.

\input{FIGURES/fig_intro}

Although these studies have shown a certain level of performance in summary evaluation, they still exhibit significant deficiencies in accuracy.
Additionally, they substantially lack in interpretability, an area crucial for further development in the field of summarization factual consistency detection.
% Also, previous studies have asserted that Natural Language Inference (NLI) models are ineffective for detecting factual consistency in summarization~\cite{falke-etal-2019-ranking, kryscinski-etal-2020-evaluating}.
As shown in Figure~\ref{fig:af}, sentence level evaluation often fails to check the details of the various facts in each sentence, resulting in lower accuracy and lower interpretability.
Furthermore, we find that pair-wise single sentence level evaluation is vulnerable to summary evaluation that requires multi-sentence reasoning. In addition, expressions such as pronouns in sentences can lead the NLI system to make incorrect judgments in single sentence level evaluation.

In this paper, we propose an interpretable summarization factual inconsistency detection system, \text{\metric}, which overcomes the issues of previous sentence level NLI-based evaluation.
%The overall pipeline of the system is illustrated in Figure~\ref{fig:overview}.
As in Figure~\ref{fig:overview}, \text{\metric} first resolves coreferences in both the source document and the generated summary.
Subsequently, we decompose this coreference resolved summary into \textit{atomic facts}, which is an approach that zooms in the summary.
This \textit{atomic fact} can be considered a more fine-grained information unit embedded within the text than a sentence at a broad level.
%Figure~\ref{fig:af} illustrates examples of \textit{atomic facts}, where a single sentence from the summary can be segmented into two or more distinct units of information.
As in the \textit{atomic fact} examples in Figure~\ref{fig:af}, a single sentence from the summary can be segmented into two or more distinct units of information.
%To address the shortcomings of the sentence level evaluation, we introduce an \textit{atomic fact} level evaluation.
This approach allows for a more detailed analysis of textual information, which is crucial for evaluating the factuality of generated text.
%After decomposing the summary into \textit{atomic facts}, we compare each \textit{atomic fact} against the source document using an NLI model to obtain scores of each atomic fact.
Using these \textit{atomic facts}, we check the consistency of each \textit{atomic fact} against the source document using an NLI model.
As highlighted in Figure~\ref{fig:af}, factual inconsistencies that cannot be detected at the sentence level can be identified through evaluation at this atomic fact level with higher interpretability.
%Following this, for certain \textit{atomic facts} that receive particularly low scores, we propose to expand the granularity of the source document to refine the comparison for the cases that require multi-sentences to check the factuality.
%Also, to check the factuality of certain atomic facts that require multi-sentence level reasoning, we propose a granularity expansion method that can adaptively increase the number of context sentences in the detection stage.
Also, we propose a granularity expansion method that can adaptively increase the number of context sentences when verifying the consistency of each atomic fact. Through this way of zooming out the document, we efficiently check the consistency of certain atomic facts that require multi-sentence level reasoning.
% We get the lowest score among these \textit{atomic facts} to compute the final score of each summary.
% This \textit{atomic fact} that received the minimum score demonstrates a high level of interpretability in FaCED by indicating whether any information in the summary, such as \textit{"Emmanuel Adebayor is 27 years old."} from Figure~\ref{fig:af}, is factually consistent or not.
% 여기 method 더 자세하게

Experimental results show that our proposed system \text{\metric} achieves state-of-the-art performance on the \textsc{AggreFact}~\cite{tang-etal-2023-understanding} benchmark dataset.
\text{\metric} exhibits high interpretability by utilizing \textit{atomic facts}.
Furthermore, We have tested on various LLMs to implement atomic fact generation task and identified the best model suited for this task.
Additionally, our analysis shows that flexibly increasing the granularity choice of the source document significantly enhances accuracy.
% 뒤에 각종 분석 내용 요약해서

%% file: FIGURES/fig_intro.tex
\begin{figure}[!t]
\centering
\includegraphics[width=\linewidth]{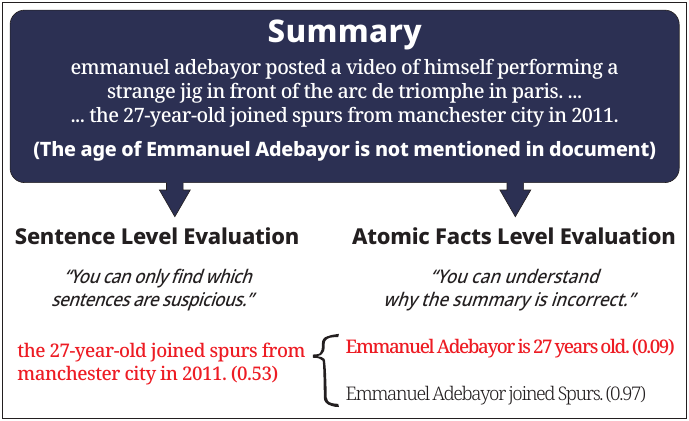}
\caption{Comparison between sentence level evaluation and atomic facts level evaluation. The numbers in parentheses represent the maximum NLI entailment scores obtained by comparing each sentence and atomic fact with the source document on a sentence-wise basis.}
\label{fig:af}
% \vspace{-5mm}
\end{figure}

%% file: 2_relatedwork.tex
\section{Related Work}
\input{FIGURES/fig_pipeline}
\paragraph{Summarization Factual Consistency Evaluation}
A multitude of metrics designed to evaluate summarization factual consistency are currently being refined by leveraging NLP pipelines originally developed for disparate tasks, including QA-based evaluation, parsing-based methods, LLM-based prompting, and NLI-based metrics.

QA-based methods involve two steps of question generation (QG) and question answering(QA).
While~\textsc{FEQA}~\cite{durmus-etal-2020-feqa} generate questions with the summary as the source, \textsc{QuestEval}~\cite{scialom-etal-2021-questeval} and \textsc{QAFactEval}~\cite{fabbri-etal-2022-qafacteval} generate questions with both the summary and the document.

Parsing-based methods discover relationships by employing syntactic parsing process, thereafter calculating the proportion of summary-derived relations that align with those extracted from source documents.
~\citet{10.1145/3292500.3330955} extract relation tuples for the evaluation.
DAE~\cite{goyal-durrett-2020-evaluating, goyal-durrett-2021-annotating} propose utilizing a dependency arc between the entities and the relationship.

There is a growing trend for using LLMs like ChatGPT~\cite{OpenAI_ChatGPT} and GPT-4~\cite{openai2023gpt4} on summarization factual consistency checking~\cite{luo2023chatgpt, chen2023evaluating, wang-etal-2023-chatgpt, gekhman-etal-2023-trueteacher, yang2024sifid}.
Initially,~\citet{luo2023chatgpt} explores ChatGPT's ability in evaluating factual consistency for text summarization with zero-shot prompting.
~\citet{yang2024sifid} extend the work by excluding irrelevant sentences from both documents before providing prompts to GPT-4.

\textsc{SummaC}~\cite{laban-etal-2022-summac} re-visit NLI-based models and granularity choice for inconsistency detection in summarization.
~\textsc{AlignScore}~\cite{zha-etal-2023-alignscore} develops an alignment system, incorporating a summarization consistency checking metric and an NLI model, which has been trained across a diverse array of tasks that can be \textit{aligned} with NLI.
The recently proposed method, FENICE~\cite{scire-etal-2024-fenice}, also aligns decomposed \textit{atomic facts} with several document sentences, but it lacks interpretability on summary side.
Our proposed system, \text{\metric}, is also based on NLI.
However, unlike the aforementioned systems, which mostly compare the summary at the sentence level, \text{\metric} conducts comparisons at a more fine-grained atomic fact level with high interpretability.

\paragraph{Atomic Facts Generation}
To the best of our knowledge, \citet{van-halteren-teufel-2003-examining} pioneered the introduction of an atomic information unit, named a \textit{factoid}, within the field of summarization evaluation.
Building on this foundational work, ~\citet{nenkova-passonneau-2004-evaluating} proposed the Pyramid method, a manual evaluation protocol for summarization that employs \textit{Summarization Content Units} (SCUs), also referred to as \textit{Semantic Content Units}.
This innovative approach has inspired a significant body of subsequent research~\cite{5eccd6e5254b425493c7acd2d642daad, shapira-etal-2019-crowdsourcing, gao-etal-2019-automated, bhandari-etal-2020-evaluating, zhang-bansal-2021-finding}.
\citet{liu-etal-2023-revisiting} referred to these elementary information units as \textit{Atomic Content Unit}, or \textit{Atomic Facts}.
However, the realm of these investigations is primarily concentrated on assessing summarization itself via the examination of \textit{atomic facts} crafted by human annotators\footnote{We note that \citet{zhang-bansal-2021-finding} generated SCUs with semantic role labeling.}.

In the scope of hallucination detection and fact verification for text generated by models, there has been a recent initiative to employ LLMs to create \textit{atomic facts}.
\textsc{FActScore}~\cite{min-etal-2023-factscore} utilize InstructGPT~\cite{ouyang2022training} for the creation of \textit{atomic facts}.
Following this work, \textsc{FacTool}~\cite{chern2023factool} introduce a fact verification pipeline that leverages fine-grained information units generated by ChatGPT, referred to as \textit{claims}.
In this study, we present a novel method \text{\metric} leveraging atomic semantic unit, from now on called \textit{atomic fact}, in the domain of summarization factual inconsistency detection.

%% file: FIGURES/fig_pipeline.tex
\begin{figure*}[t]
\centering
\includegraphics[width=1.0\linewidth]{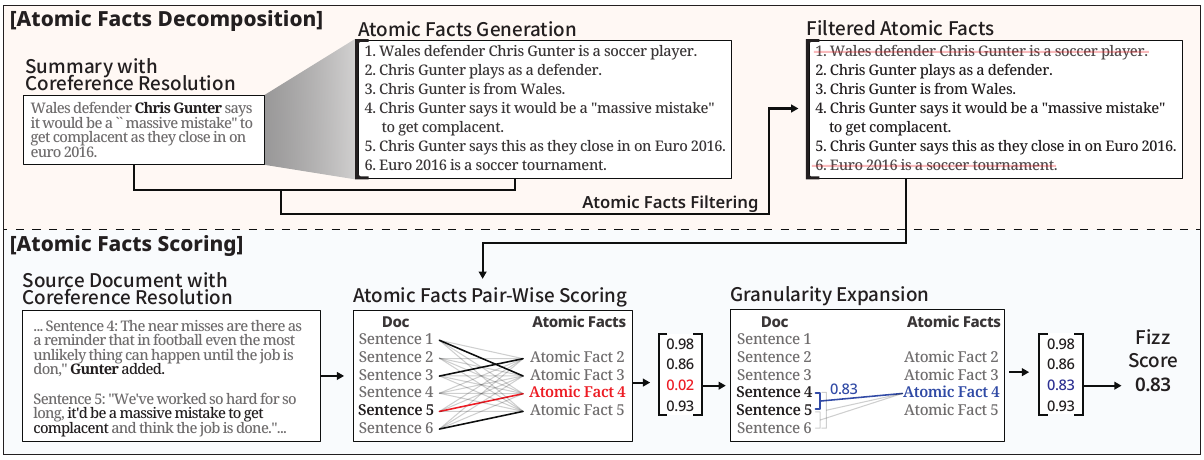}
\caption{Overall flow of \text{\metric}. The pipeline begins by applying coreference resolution to both the summary and the document. Atomic facts are then decomposed from the summary using an LLM. These atomic facts are filtered and subsequently scored against the document. The scores are refined through granularity expansion. The ultimate score is defined by choosing the minimum score.}
\label{fig:overview}
%\vspace{-5mm}
\end{figure*}

%% file: 3_method.tex
\section{\text{\metric}}
The overall flow of our proposed system \text{\metric} is presented in Figure~\ref{fig:overview}.
Our method first begins with the application of a coreference resolution model to a given $(document, summary)$ pair, resulting in a new pair of texts $(document, summary)$ where coreferences have been resolved (Section~\ref{sec:coref_resolve}).
%Following this, we proceed to generate \textit{atomic facts} from the coreference-resolved summary leveraging LLMs, with this procedure being executed at the granularity of sentence level (Section~\ref{sec:atomicfact_decomp}).
Following this, we proceed to generate \textit{atomic facts} from the coreference-resolved summary leveraging LLMs as a zooming-in approach for the summary (Section~\ref{sec:atomicfact_decomp}). 
Using the generated atomic facts, we compute the score of each \textit{atomic fact} with the NLI system (Section~\ref{sec:nli_score}).
%Finally, to precisely compute the score, we also propose granularity expansion method, which is a way of zooming out the documents, within the source document to handle the summaries that contain high abstractiveness.
Finally, we propose a granularity expansion method, which is a way of zooming out the documents, to compute the score for the summaries that contain high abstractiveness more accurately.

% -----------------------------------------------------------------------------
% Coreference Resolution
% -----------------------------------------------------------------------------
\subsection{Coreference Resolution}
\label{sec:coref_resolve}
% coref resolution을 document, summary에 둘 다 우선적으로 적용한다.
% In this work, we prioritize conducting coreference resolution on both the document and summary texts. The rationale for this approach is twofold. Firstly, pronoun resolution is applied to summaries because, rather than feeding the entire summary as a prompt into an LM for atomic fact generation, we extract atomic facts on a sentence-by-sentence basis from the summary. Secondly, the necessity for pronoun resolution in the document originates in the characteristics of NLI models. NLI models, when presented with premises and hypotheses containing the same content, may fail to recognize entailment if the premise includes pronouns while the hypothesis contains explicit entity names. This discrepancy highlights the importance of resolving coreferences in documents to align with the explicit entity names in hypotheses, thereby enhancing the entailment recognition capabilities of NLI models.
To enhance the entailment recognition capabilities of NLI models, \text{\metric} first conducts centered around coreference resolution in both document and summary texts.
The motivation behind this approach is driven by the inherent limitations observed in NLI models when processing texts with pronouns.
Specifically, we find that NLI models tend to struggle with recognizing entailment when presented with \textit{premises} and \textit{hypotheses} that contain the same content but differ in their use of pronouns and explicit entity names.
To address this challenge, \text{\metric} employs pronoun resolution in summaries by analyzing them on a sentence-by-sentence basis to extract atomic facts.
This strategy not only facilitates a more granular understanding of the summary content but also avoids the limited context length in LLMs. 

Furthermore, applying pronoun resolution to the document text ensures that the entities are explicitly named, aligning the \textit{premise} more closely with the \textit{hypothesis}.
By resolving coreferences in both documents and summaries, our approach aims to bridge the gap between pronoun use and explicit entity naming, thereby improving the performance of NLI models in entailment tasks.
This dual focus on both document and summary texts underscores the comprehensive nature of our strategy to bolster the accuracy and reliability of NLI models in handling a variety of linguistic expressions.
% 관련 citation 뭐 없나..?

Formally, given a document $D$ and its summary $S$, we define coreference resolution as $f_{\text{coref}}$, which makes:
\begin{equation}
\label{eq:coref_resolv}
D^\prime = f_{\text{coref}}(D),\ \ S^\prime = f_{\text{coref}}(S)
\end{equation}
where $D^\prime$ and $S^\prime$ are coreference resolved texts of $D$ and $S$, respectively.

% -----------------------------------------------------------------------------
% Atomic Fact Decomposition
% -----------------------------------------------------------------------------
\subsection{Atomic Facts Decomposition}
\label{sec:atomicfact_decomp}
\paragraph{Atomic Facts Generation}
As demonstrated in Figure~\ref{fig:af}, sentence level evaluation of summaries can often yield inaccurate results.
Therefore, we propose a method that evaluates the factuality of summaries at a more fine-grained level, specifically at the level of \textit{atomic facts} as exemplified in Figure~\ref{fig:overview}.
By employing \textit{atomic facts}, which are highly detailed units of information, \text{\metric} considerably enhances interpretability.

% Formulating a definition for \textit{atomic fact} could be inherently impossible~\cite{liu-etal-2023-revisiting}. 
The definition of an \textit{atomic fact} differs across studies, primarily due to the inherently subjective nature of this concept.
We propose our own definition of an \textit{atomic fact} that is designed to align with and complement the nature of NLI models.
Building upon~\citet{bhandari-etal-2020-evaluating}, we specify further that an \textit{atomic fact} is short and concise, containing no more than two or three entities, with person entities specifically resolved any of coreferences.

We generate atomic facts from summaries at the sentence level after resolving coreferences.
This strategy for atomic fact generation not only increases the quantity of atomic facts but also substantially augments the generated summary's pool of information.
To extract atomic facts from the summaries, we input prompts into the LLM that include both a task description and a sentence-level summary, as exemplified in Table \ref{tab:appx_prompt}.
% prompt appendix로 옮겨졌는데 그대로 유지해야할까?
This approach systematically decomposes each sentence in the summary into individual atomic facts, facilitating a comprehensive extraction and representation of information.
The coreference resolved summary $S^\prime=\{s_j^\prime\}_{j=1}^{N}$, where $s_j^\prime$ represents the $j^{th}$ sentence in $S^\prime$ and $N$ the total number of sentences in $S^\prime$, could be decomposed to a set of atomic facts $A^\prime=\{a_k^\prime\}_{k=1}^{L}$, with $L$ denotes the total number of sentences in $A^\prime$.
% \begin{equation}
% \label{eq:atomic_fact}
% \begin{aligned}
% {A_j^\prime} = f_{\text{decomp}}({s_j^\prime})_{0\le j\le N} \\
% {A_j^\prime} = \{{a_{j, k}^\prime}\}_{0\le j\le N, 0\le k\le L}\\
% {A^\prime} = \{{A_{j}^\prime}\}_{0\le j\le N}
% \end{aligned}
% \end{equation}

\paragraph{Atomic Facts Filtering}
One significant issue with atomic facts generated by LLMs is that these facts are often produced not from the content of summaries themselves but from the pretrained knowledge embedded within the LLMs.
For example, when we decompose the sentence of the summary \textit{"The mass, which has risen some 50ft above sea level, measures roughly 1,000 - 1,640ft long, and 100ft wide"}, the decomposed atomic facts contain an atomic fact \textit{"The mass is a noun"}.
Such atomic facts may not align with either the summaries or the documents and can significantly influence the scoring method described in Section~\ref{sec:nli_score}. %~\hyperref[sec:nli_score]{3.3}.
Consequently, the exclusion of these atomic facts becomes a necessary step in our process.

Hence, we utilize an NLI model to filter out incorrect atomic facts.
Our approach leverages the probabilistic distribution of the NLI model, which categorizes outcomes into three types: \textit{Entailment} (E), \textit{Contradiction} (C), and \textit{Neutral} (N).
In the filtering process, we set the summary $S^\prime$ as the \textit{premise}, and the atomic fact $A^\prime$ as the \textit{hypothesis}.
We filter out atomic facts that exhibit exceptionally low \textit{entailment} scores.
We outline the detailed procedure of the atomic facts filtering process in Algorithm \ref{alg:atomic_fact_filtering}.
% Sentence pair-wise of $\{s_j^\prime\}_{j=0}^{N}$ and $\{a_{j,k}^\prime\}_{k=0}^{L}$ through NLI model results in $(e_{j, k}, c_{j, k}, n_{j, k})$.
% We retain atomic facts $A^*$ for which $e_{j, k}\ge c_{j, k}$ and $e_{j, k}\ge n_{j, k}$:
% \begin{equation}
% {A^*} = \{{a_{j,k}^\prime}\}_{0\le j\le N, 0\le k\le L, e_{j, k}\ge c_{j, k}, e_{j, k}\ge n_{j, k}}
% \end{equation}
\begin{algorithm}[t]
    \small
    \caption{Filtering Out Incorrect Atomic Facts}
    \textbf{Input}: An NLI model $\mathcal{M}$; coreference resolved summary $S^\prime=\{s_j^\prime\}_{j=1}^{N}$; decomposed atomic facts $A^\prime=\{a_{k}^\prime\}_{k=1}^{L}$.
    \textbf{Initialize}: set $A_{filtered}=\phi$
    \begin{algorithmic}[1]
        \For{$k = 1, 2, \ldots, L$}
            \For{$j = 1, 2, \ldots, N$}
            \State $(e_{j,k}, c_{j,k}, n_{j,k}) \leftarrow \mathcal{M}(s_j^\prime, a_k^\prime)$
            \If {$\max(e_{j,k}, c_{j,k}, n_{j,k})$ is $e_{j,k}$}
            \State Append $a_k^\prime$ to $A_{filtered}$.
            % \State $A_{filtered}$.append($a_k^\prime$)
            \EndIf
            \EndFor
        \EndFor
    \end{algorithmic}
    \textbf{Output}: A set of atomic facts $A_{filtered}$.
    \label{alg:atomic_fact_filtering}
\end{algorithm}
%\vspace{-5mm}
% -----------------------------------------------------------------------------
% Scoring
% -----------------------------------------------------------------------------
\subsection{Atomic Facts Scoring}
\label{sec:nli_score}
\paragraph{Atomic Facts Pair-Wise Scoring}
To compute the score for each atomic fact of the summaries, \text{\metric} first decomposes the coreference resolved document into sentences.
We split the document $D^\prime$ into M sentences and the filtered atomic facts $A_{filtered}$ into N sentences, formulating $D^\prime=\{d_i^\prime\}_{i=1}^{M}$ and $A_{filtered}=\{a_k\}_{k=1}^{L}$, respectively.
We use each $(d_i, a_k)$ as an input for an NLI model, positioning the generated atomic fact as the \textit{hypothesis} and the sentence of the document as the \textit{premise}.

%Consistent with findings from previous research~\cite{glover-etal-2022-revisiting}, our experiments indicate that employing exclusively the \textit{entailment} score leads to enhanced outcomes.
Finally, we assign scores to each atomic fact based on the \texttt{maximum} \textit{entailment} score obtained through comparison with every sentence in the document.
The atomic fact \textit{entailment} scores $E=\{e_{i, k}\}$, where $1\le i\le M$ and $1\le k\le L$, are computed to a vector $\mathbf{T}$:

\begin{equation}
\begin{gathered}
\mathbf{t}_{k} = \max_{1\le i\le M}e_{i,k} \\
\mathbf{T} = \{\mathbf{t}_1, \ldots, \mathbf{t}_L\}
\end{gathered}
\end{equation}

\paragraph{Adaptive Granularity Expansion}
Summaries generated by \textit{abstractive} summarization systems contain a high degree of \textit{abstractiveness}.
This \textit{abstractiveness} occurs when content spread across multiple sentences in the document is condensed into one or two sentences in the summary.
% To accurately detect factual inconsistencies within such summaries, it is necessary to examine multiple sentences across the source document.
To accurately detect factual inconsistencies within such summaries, it is necessary to zoom out and examine multiple sentences across the source document.
Furthermore, several studies have demonstrated that considering multiple sentences from the document leads to better accuracy~\cite{laban-etal-2022-summac, glover-etal-2022-revisiting}.

We aim to identify scores where $\max(e_k, c_k, n_k)$ is \textit{not} $e_k$ from the $\mathbf{T}$.
For atomic facts associated with these scores, we further increase the granularity of the document and perform computation once again.
We incrementally increase the granularity starting from the document sentence $d_i$ that contributed to each identified score, limiting the granularity at a maximum of three sentences ($d_{i-1}$ + $d_i$, $d_{i}$ + $d_{i+1}$, $d_{i-2}$ + $d_{i-1}$ + $d_i$, $d_{i}$ + $d_{i+1}$ + $d_{i+2}$, $d_{i-1}$ + $d_i$ + $d_{i+1}$).
Subsequently, we re-calculate the scores within this expanded context and replace the original scores with the \texttt{maximum} value observed among the re-calculated scores and the original.
As a result, the vector $\mathbf{T}$ is transformed into $\mathbf{T^\ast}$ as certain scores are replaced by new scores.
Detailed information on this procedure is provided in Algorithm~\ref{alg:doc_ge}.

\begin{algorithm}[t]
    \small
    \caption{Scoring with Document Granularity Expansion}
    \textbf{Input}: An NLI model $\mathcal{M}$; coreference resolved document $D^\prime=\{d_i^\prime\}_{i=1}^{M}$; decomposed atomic facts $A^\prime=\{a_{k}^\prime\}_{k=1}^{L}$.
    
    \textbf{Initialize}:
    $\mathbf{T^\ast}=\phi$; Max granularity size $gran = 3$.
    \begin{algorithmic}[1]
        \State Define $\mathcal{C}(D, g)$ = list of subsets of $D$ with size of $g$.
        \State Define $\mathcal{F}(\mathcal{C}(D, g))$ which returns whether $\mathcal{C}(D, g)$ is a consecutive list.
        \State Define $\mathcal{D}(\mathcal{C}(D, g))$ = list of document sentences in index list in $\mathcal{C}(D, g)$.
        \For{$k = 1, 2, \ldots, L$}
        \State set $E = \phi$
            \For{$i = 1, 2, \ldots, M$}
            \State $(e_{i,k}, c_{i,k}, n_{i,k}) \leftarrow \mathcal{M}(d_i^\prime, a_k^\prime)$
            \State Append $e_{i,k}$ to $E$.
            \EndFor
        \State $m_{idx} = E.index(\max(E))$
        \If {$\max(e_{i,k}, c_{i,k}, n_{i,k})$ is \textbf{not} $e_{i,k}$}
        \State set $D_{idx}=[0, \ldots, M-1]$
        \State set $D_{expanded}=\phi$
            \For{$g = 1, 2, \ldots, gran+1$}
            \If {$m_{idx}$ in $\mathcal{C}(D_{idx}, g)$ and $\mathcal{F}(\mathcal{C}(D_{idx}, g))$}
            \State Extend $\mathcal{C}(D_{idx}, g)$ to $D_{expanded}$.
            \EndIf
            \EndFor
            \State set $E_{expanded} = \phi$
            \For{$d_{expanded}\,\in\,\mathcal{D}(D_{expanded})$}
            \State $(e, c, n) \leftarrow \mathcal{M}(d_{expanded}, a_k^\prime)$
            \State Append $e$ to $E_{expanded}$. 
            \EndFor
        \State Append $\max(E_{expanded})$ to $\mathbf{T^\ast}$.
        \Else
        \State Append $e_{i,k}$ to $\mathbf{T^\ast}$.
        \EndIf
        \EndFor
    \end{algorithmic}
    \textbf{Output}: vector $\mathbf{T^\ast}$ with maximum entailment scores from each atomic fact. 
    \label{alg:doc_ge}
    %\vspace{-5mm}
\end{algorithm}

The final score is then determined by the \texttt{minimum} score within vector $\mathbf{T^\ast}$, enabling a highly interpretable evaluation:
% The final score is then determined by the \texttt{minimum} score within vector $\mathbf{T^\ast}$, which makes it much easier to notice whether the summary contains inconsistency:
\begin{equation}
\textit{\metric}\,\,score = \min(\mathbf{T^\ast})
\end{equation}

%% file: 4_experiments.tex
\section{Experiments}
\label{sec:experiments}
\subsection{Experimental Setups}
In our experiments, we leverage MT5~\cite{bohnet-etal-2023-coreference} for coreference resolution, which returns with the identification of clusters referring to the same entities.
With these clusters, we further implement rule-based pronoun substitution strategies to generate coreference resolved texts.
For atomic fact decomposition, the Orca-2 model~\cite{mitra2023orca} is utilized.
Additionally, this work adopts the same off-the-shelf NLI model as implemented in \textsc{SummaC} (See Appendix~\ref{sec:exp_details_nli} for more details).

\subsection{Benchmark Datasets}
We use \textsc{AggreFact}~\cite{tang-etal-2023-understanding} benchmark dataset, a comprehensive aggregation of 9 leading summary factual consistency detection datasets currently available.
\textsc{AggreFact} is stratified into three distinct splits, namely \textsc{FtSota}, \textsc{EXformer}, and \textsc{Old}, with each split containing its own validation and test sets.
We standardize the evaluation as a binary classification and choose the best threshold from the validation set following SummaC.
Finally, we apply this threshold to the test set and report the balanced accuracy score, considering the imbalance in the dataset.

\subsection{Baselines}
We adopt all of the baselines of \textsc{AggreFact} dataset: DAE~\cite{goyal-durrett-2020-evaluating, goyal-durrett-2021-annotating}, QuestEval~\cite{scialom-etal-2021-questeval}, SummaC-ZS and SummaC-Conv~\cite{laban-etal-2022-summac}, QAFactEval~\cite{fabbri-etal-2022-qafacteval}, ChatGPT-ZS and ChatGPT-CoT~\cite{luo2023chatgpt}, ChatGPT-DA and ChatGPT-Star~\cite{wang-etal-2023-chatgpt}.
Also, we report the results with AlignScore~\cite{zha-etal-2023-alignscore}, which is a recently introduced system for checking the factual consistency of summaries based on NLI.
\input{TABLES/aggrefact_ftsota}
Additionally, we incorporate \textsc{FActScore}~\cite{min-etal-2023-factscore} and \textsc{FacTool}~\cite{chern2023factool} in our baselines.
These methods decompose generated texts into \textit{atomic facts} and then retrieve corresponding entries from a given knowledge base, such as Wikipedia, to evaluate the factuality of the generated context.
For the purpose of verification, we assume the availability of this knowledge base, which we use as the source document to assess summary factual consistency.
In \textsc{FActScore}, we employ a \textbf{No-context LM} for factual verification.
This approach operates on a QA basis, assessing whether \textit{atomic facts} are true or false with respect to the source document.
In \textsc{FacTool}, we utilize a \textbf{Knowledge-based QA} approach.
% This also follows a QA format but incorporates the CoT method, where the LLM evaluates if \textit{claims} are true or false relative to the source document\footnote{Details of the experiments are provided in Appendix~\ref{sec:exp_details}}.
This also follows a QA format but incorporates the CoT method, where the LLM evaluates if \textit{claims} are true or false relative to the source document.
Details of the experiments are provided in Appendix~\ref{sec:exp_details_baselines}.

\input{TABLES/aggrefact_total}
\subsection{Results}
We present the performance outcomes obtained by applying each metric to the \textsc{AggreFact} benchmark dataset in Table~\ref{tab:aggrefact}.
We show the performance of three versions of our proposed metric: \text{\metric}, its without granularity expanded version, \text{\metric$_{w/o\,\,GE}$}, and its without atomic facts version, \text{\metric$_{w/o\,\,AF}$}.
The complete results for \textsc{AggreFact-Cnn} and \textsc{AggreFact-XSum} are displayed in Table~\ref{tab:aggrefact}.
\text{\metric} demonstrates the highest average performance, followed by \text{\metric$_{w/o\,\,GE}$} and \text{\metric$_{w/o\,\,AF}$}.

Additionally, we provide results for a single-threshold approach on \textsc{AggreFact-FtSota} split as in~\citet{tang-etal-2023-understanding}.
We list the best threshold findings for the \textsc{AggreFact-Cnn-FtSota} and \textsc{AggreFact-XSum-FtSota} splits, with corresponding binary classification balanced accuracy scores in Table \ref{tab:aggrefact_ftsota}.
In this setting, \text{\metric} achieves the highest average performance, with \text{\metric$_{w/o\,\,GE}$} coming in second.
Both metrics perform exceptionally well on the \textsc{Cnn} split.
Furthermore, the granularity expansion in \text{\metric} leads to notably higher performance improvements on the \textsc{XSum} split.

Both \textsc{FActScore} and \textsc{FacTool} have demonstrate scores that are comparable to or exceed those of ChatGPT-based metrics.
It appears that decomposing summaries into atomic facts and comparing them with the source document is more effective than performing factuality checking on the entire summary.
However, metrics based on ChatGPT inherently face disadvantages compared to other metrics, which can be tuned by adjusting thresholds; such tuning is unnecessary for ChatGPT-based metrics.
This distinction may limit the effectiveness of ChatGPT-based evaluations in some contexts.
\input{TABLES/analysis_llms}
\input{TABLES/analysis_af}

\subsection{Analysis}
\paragraph{LLMs used for Atomic Facts Decomposition}
\label{sec:llm_analysis}
To investigate the most suitable LLMs for generating atomic facts, we evaluate the generation of atomic facts using various LLMs, including \texttt{gpt-3.5-turbo}, \texttt{gpt-3.5-turbo-instruct}, and other 7B models such as Zephyr~\cite{tunstall2023zephyr} and Mistral~\cite{jiang2023mistral}.
The results, documented in Table \ref{tab:analysis_llms}, demonstrate that while the atomic facts generated by \texttt{gpt-3.5-turbo} and \texttt{gpt-3.5-turbo-instruct} generally perform better compared to other metrics, they are still inferior to those produced by Orca-2.
The performance drop associated with the \texttt{gpt} series suggests a noteworthy observation.
We explain that this discrepancy is due to the length of the atomic facts.
As shown in Table~\ref{tab:analysis_llms}, which includes the average token length of atomic facts after the filtering process per summary, there is a clear inverse relationship between the number of tokens in an atomic fact and its average performance.
Longer atomic facts tend to contain more entities and are less concise.
Such sentences are less suitable as \textit{hypotheses} when compared sentence-wise using NLI models.
Furthermore, the sensitivity of using the minimum atomic fact scores as the final score exacerbates the challenge, making it difficult to achieve desired outcomes with lengthy sentences. In contrast, other 7B models such as LLaMa~\cite{Touvron2023LLaMAOA} show limitations in adhering to instructions for atomic fact decomposition.
Details of the model usage are provided in Appendix~\ref{sec:exp_details_llms}.

\input{FIGURES/fig_ge}
% 추가본
In previous studies~\cite{zhang-bansal-2021-finding, chern2023factool, scire-etal-2024-fenice}, the evaluation of the quality and the completeness of the LLM generated atomic facts focuses solely on content similarity (\ie{, ROUGE-1}) with human-written atomic facts.
However, we consider content similarity evaluation to be insufficient and added two additional factors: 1) Average token length in atomic facts and 2) Average number of atomic facts.
In Table~\ref{tab:analysis_llms}, we demonstrate the correlation between the average token length of atomic facts and overall performance.
Building on this, we now analyze the token length of both human-written and generated atomic facts.
Additionally, since the content similarity metric does not take into account the number of atomic facts, we also include the average number of atomic facts in our results.
We report the comparative analysis of the LLM generated atomic facts against human-written atomic facts in Table~\ref{tab:analysis_af}.
The experiments were implemented using the RoSE~\cite{liu-etal-2023-revisiting} dataset, which includes 2,500 summaries and their corresponding human-written atomic facts.
As shown in the experimental results, \texttt{gpt-3.5-turbo} demonstrates the highest capability by achieving the top score in content similarity.
However, it shows a significant difference in the number of atomic facts and the number of tokens in atomic facts.
In contrast, Mistral scores lower in content similarity but exhibits higher human correlation in the number of atomic facts and token lengths.
The model that achieves the highest human correlation in both the number of atomic facts and token lengths is Orca-2, which shows the best performance among LLMs as in Table~\ref{tab:analysis_llms}.
These findings suggest that while content similarity is important, the number of atomic facts and token lengths are equally critical factors to consider.
Details on computing content similarity are provided in Appendix~\ref{sec:exp_details_ccs}.
% 추가본

\input{TABLES/analysis_granularity}
\paragraph{Sizes of Granularity Expansion}
\label{sec:ge_analysis}
As underscored in Section~\ref{sec:nli_score}, accurately evaluating the factual consistency of \textit{abstractive} summaries necessitates an expansion of document granularity.
This requirement stems from the observation that a single sentence within a summary may incorporate content from multiple sentences within the document.
Illustrative of this point, Figure \ref{fig:GE} highlights that segmenting conversational dialogues into discrete sentences can lead to a loss of contextual clarity, where the synthesis of various segmented sentences is required for an accurate interpretation.

\textsc{SummaC} present experimental results across different granularity choices, categorizing document granularity into a sentence, two sentences, paragraph, and full document levels.
However, adjusting document granularity in such a manner reduces interpretability and undermines result reliability.
Our approach is to adaptively increase granularity only for atomic facts where the entailment score significantly decreases.

Table \ref{tab:analysis_granularity} presents the outcomes associated with varying granularity sizes in adaptive granularity expansion.
The experimental findings reveal a consistent improvement in average performance with increasing granularity, particularly for summaries derived from XSum~\cite{narayan-etal-2018-dont}.
This significant performance boost can be attributed to the inherently \textit{abstractive} nature of XSum-based summaries. 

Despite the increase in average score for the maximum of four sentences, the scores for CNN summaries actually declined.
Additionally, we observe that computational costs rose with increasing granularity.
Hence, we determined that the maximum of three sentences represents the best trade-off between computational cost and performance.
Details on granularity expansion condition choice are provided in Appendix~\ref{sec:exp_details_ge}.
\input{TABLES/analysis_coref}

\paragraph{Effectiveness of Coreference Resolution}
\label{sec:cr_analysis}
In the application of NLI models for comparing \textit{premises} with \textit{hypotheses}, the significance of coreference resolution cannot be overstated.
As outlined in Section~\ref{sec:coref_resolve}, failure to resolve pronouns in the \textit{premise} significantly hinders the attainment of desired outcomes.
This point is vividly illustrated in Figure \ref{fig:GE}, where the difference between document(b) and document(c) is merely the resolution of pronouns.
Yet, this seemingly minor modification leads to a stark contrast in entailment scores, with document(b) achieving a score of 0.09 compared to document(c)'s 0.83.
The discrepancy arises due to the document (\textit{premise})'s reference to "\textit{he}" not being recognized as pertaining to "\textit{Chris Gunter}", as stated in the atomic fact (\textit{hypothesis}).

Moreover, Table \ref{tab:analysis_coref} presents more granular experimental results on the impact of coreference resolution.
We implemented experiments to evaluate the impact of coreference resolution on both documents and atomic facts.
Our investigation included scenarios where coreference resolution was applied and cases where it was not.
We show that texts with resolved coreferences, whether they be atomic facts or documents, consistently outperform those without resolution.
Notably, there is a marked improvement in performance on datasets based on CNN~\cite{hermann2015teaching} summaries compared to those based on XSum summaries.
This is likely due to the \textit{extractive} nature of CNN-based summaries, as opposed to the more \textit{abstractive} summaries derived from XSum.
Details on coreference resolution usage are provided in Appendix~\ref{sec:exp_details_cr}.

\input{FIGURES/fig_case}
\paragraph{Failure Case Study}
We analyze the drawbacks of decomposing summaries into atomic facts in the summary factual consistency checking task, through the main example in Figure \ref{fig:case}, which compares the drawbacks of analyzing atomic facts versus sentences.
When comparisons are made at the sentence level, a sentence can be correctly judged as entailing the content of a document.
Conversely, when breaking down the content into atomic facts, the fact \textit{"The tweet was about a rocket landing."} receives a maximum entailment score of only 0.33.
This particular atomic fact remains even after undergoing the filtering process.
As a result, a summary that is factually consistent may be erroneously classified as factually inconsistent due to the analysis of this single atomic fact.

%% file: TABLES/aggrefact_ftsota.tex
\begin{table}[!t]
    \centering
    \small
    \resizebox{\linewidth}{!}{
        \begin{tabular}{lcc|c}
            \toprule
            & \begin{tabular}[c]{@{}c@{}}\textbf{\textsc{AggreFact-}}\\\textbf{\textsc{Cnn-FtSota}}\end{tabular} & \begin{tabular}[c]{@{}c@{}}\textbf{\textsc{AggreFact-}}\\\textbf{\textsc{XSum-FtSota}}\end{tabular} & \begin{tabular}[c]{@{}c@{}}\textbf{\textsc{Avg}}\end{tabular} \\
            \midrule
            DAE          & 65.4$\pm$4.4 & \textbf{70.2}$\pm$2.3 & 67.8 \\
            QuestEval    & 70.2$\pm$3.2 & 59.5$\pm$2.7 & 64.9 \\
            SummaC-ZS    & 64.0$\pm$3.8 & 56.4$\pm$1.2 & 60.2 \\
            SummaC-Conv  & 61.0$\pm$3.9 & 65.0$\pm$2.2 & 63.0 \\
            QAFactEval   & 67.8$\pm$4.1 & 63.9$\pm$2.4 & 65.9 \\
            AlignScore   & 62.5$\pm$3.3 & 69.6$\pm$1.7 & 66.1 \\
            \midrule
            ChatGPT-ZS   & 56.3$\pm$2.9 & 62.7$\pm$1.7 & 59.5 \\
            ChatGPT-COT  & 52.5$\pm$3.3 & 55.9$\pm$2.1 & 54.2 \\
            ChatGPT-DA   & 53.7$\pm$3.5 & 54.9$\pm$1.9 & 54.3 \\
            ChatGPT-Star & 56.3$\pm$3.1 & 57.8$\pm$0.2 & 57.1 \\
            \midrule
            FactScore    & 60.8$\pm$3.2 & 68.0$\pm$2.0 & 64.4 \\
            FacTool      & 49.3$\pm$3.5 & 59.0$\pm$2.0 & 54.2 \\
            \midrule
            % \text{\metric$_{g}$}
            \text{\metric} \textbf(Ours) & \textbf{72.6}$\pm$3.0 & 69.3$\pm$1.9  & \textbf{71.0} \\
            \quad${w/o\,\,GE}$ & \underline{72.2}$\pm$2.8 & 66.3$\pm$1.9  & \underline{69.3} \\
            \quad${w/o\,\,Filtering}$ & 64.7$\pm$3.3 & \underline{70.0}$\pm$1.8 & 67.4 \\
            \quad${w/o\,\,AF}$ & 63.6$\pm$2.9 & 65.8$\pm$2.0 & 64.7 \\
            \bottomrule
        \end{tabular}
    }
    \caption{Balanced accuracy using a single threshold with 95\% confidence intervals on the \textsc{AggreFact-FtSota} split dataset. Highest performance is highlited in \textbf{bold}, and the second highest is \underline{underlined}.} 
\label{tab:aggrefact_ftsota}
%\vspace{-3mm}
\end{table}

%% file: TABLES/aggrefact_total.tex
\begin{table}[!t]
\setlength{\abovecaptionskip}{0.3cm}
\setlength{\belowcaptionskip}{-0.1cm}
\resizebox{\linewidth}{!}{
    \centering
    \small
    \setlength\tabcolsep{3pt}  % Reduce column padding
        \begin{tabular}{lccc|ccc|c}
            \toprule
            & \multicolumn{3}{c|}{\textbf{\textsc{AggreFact-Cnn}}} & \multicolumn{3}{c|}{\textbf{\textsc{AggreFact-XSum}}} & \multicolumn{1}{c}{} \\
            & \textbf{\textsc{FtSota}} & \textbf{\textsc{EXf}} & \textbf{\textsc{Old}} & \textbf{\textsc{FtSota}} & \textbf{\textsc{EXf}} & \textbf{\textsc{Old}} & \textbf{\textsc{Avg}} \\
            % & \begin{tabular}[c]{@{}c@{}}\textbf{\textsc{AggreFact-}}\end{tabular} & \begin{tabular}[c]{@{}c@{}}\textbf{\textsc{AggreFact-}}\end{tabular} & \begin{tabular}[c]{@{}c@{}}\textbf{\textsc{Avg}}\end{tabular} \\
            \midrule
            Baseline & 50.0 & 50.0 & 50.0 & 50.0 & 50.0 & 50.0 & 50.0 \\
            \midrule
            DAE* & 59.4 & 67.9 & 69.7 & \textbf{73.1} & - & - & 67.5 \\
            QuestEval & 63.7 & 64.3 & 65.2 & 61.6 & 60.1 & 59.7 & 62.4 \\
            SummaC-ZS & 63.3 & \textbf{76.5} & 76.3 & 56.1 & 51.4 & 53.3 & 62.8 \\
            SummaC-Cv & 70.3 & 69.8 & 78.9 & 67.0 & 64.6 & 67.5 & 69.7 \\
            QAFactEval & 61.6 & 69.1 & \textbf{80.3} & 65.9 & 59.6 & \underline{60.5} & 66.2 \\
            AlignScore & 53.4 & \underline{73.1} & \underline{80.2} & \underline{70.2} & \textbf{80.1} & 63.7 & 70.1 \\
            \midrule
            ChatGPT-ZS & 66.2 & 64.5 & 74.3 & 62.6 & 69.2 & 60.1 & 66.2 \\
            ChatGPT-CoT & 49.7 & 60.4 & 66.7 & 56.0 & 60.9 & 50.1 & 57.3 \\
            ChatGPT-DA & 48.0 & 63.6 & 71.0 & 53.6 & 65.6 & 61.5 & 60.6 \\
            ChatGPT-Star & 55.8 & 65.8 & 71.2 & 57.7 & 70.6 & 53.8  & 62.5 \\
            \midrule
            FactScore & 69.9 & 71.6 & 73.9 & 68.0 & 63.5 & 66.8 & 69.0 \\
            FacTool & \underline{72.7} & 66.1 & 60.8 & 68.0 & 64.0 & 62.2 & 65.6 \\
            \midrule
            \text{\metric} \textbf(Ours) & \textbf{73.2} & 67.3 & 76.0 & 69.7 & \underline{72.4} & \textbf{68.5} & \textbf{71.2} \\
            %\quad${w/o\,\,GE}$ & 68.0 & 67.1 & 77.6 & 69.4 & 71.5 & \textbf{69.8} & \underline{70.6} \\
            %\quad${w/o\,\,AF}$ & 72.4 & 67.0 & \textbf{80.4} & 64.9 & 67.1 & 66.1 & 69.7 \\
            \bottomrule
        \end{tabular}
    }
    \caption{Balanced accuracy on the \textsc{AggreFact} dataset. As in~\citet{tang-etal-2023-understanding}, we omitted the results from DAE, as it was trained on the XSumFaith~\cite{goyal-durrett-2021-annotating} dataset, which includes human-annotated summaries from \textsc{EXformer} and \textsc{Old}.}
\label{tab:aggrefact}
%\vspace{-5mm}
\end{table}

%% file: TABLES/analysis_llms.tex
% \begin{table}[]
%     \centering
%     \resizebox{\linewidth}{!}{
%         \begin{tabular}{c|cc|c}
%             \toprule
%             \textbf{LLM} & \textbf{\textsc{Cnn}} & \textbf{\textsc{XSum}} & \textbf{\textsc{Avg}} \\
%             \midrule
%             \texttt{gpt-3.5-turbo-1106} & 65.9$\pm$3.3 & 64.7$\pm$1.9  & 65.3 \\
%             \midrule
%             \texttt{gpt-3.5-turbo-instruct} & 68.6$\pm$3.1 & 66.2$\pm$2.1  & 67.4 \\
%             \midrule
%             \texttt{Orca-2-7B} & \textbf{72.2}$\pm$2.7 & \textbf{66.3}$\pm$1.9  & \textbf{69.3} \\
%             \bottomrule
%         \end{tabular}
%     }
%     \caption{Experimental results of our metric with atomic facts generated by different LLMs and prompts on \textsc{AggreFact-FtSota} split. We only altered the prompt for atomic fact generation, while all other processes before the granularity expansion were conducted identically to (Metric Name).}
% \label{tab:analysis_llms}
% \end{table}

\begin{table}[]
    \centering
    \small
    \resizebox{\linewidth}{!}{
        \begin{tabular}{c|cc|c|c}
            \toprule
            \textbf{\textsc{LLM}} & \textbf{\textsc{Cnn}} & \textbf{\textsc{XSum}} & \textbf{\textsc{Avg}} & \textbf{\textsc{Avg. Token Length}}\\
            \midrule
            \texttt{Zephyr} & 65.1$\pm$3.3 & 65.2$\pm$2.0 & 65.2 & \textbf{97.6} \\
            \texttt{gpt-3.5-turbo} & 68.7$\pm$3.4 & 68.7$\pm$2.0 & 68.7 & 95.9 \\
            \texttt{gpt-3.5-turbo-instruct} & 70.7$\pm$3.1 & 67.0$\pm$1.8 & 68.9 & 90.5 \\
            \texttt{Mistral} & 70.5$\pm$3.5 & 68.7$\pm$2.1 & 69.6 & 86.5 \\
            \midrule
            \texttt{Orca-2} & \textbf{72.6}$\pm$3.0 & \textbf{69.3}$\pm$1.9  & \textbf{71.0} & 81.4 \\
            \bottomrule
        \end{tabular}
    }
    \caption{Experimental results of \text{\metric} with atomic facts generated by different LLMs using the same prompt on \textsc{AggreFact-FtSota} split. 
    \textbf{Avg. Token Length} indicates the average number of total tokens of atomic facts per summary.}
\label{tab:analysis_llms}
%\vspace{-2mm}
\end{table}

%% file: TABLES/analysis_af.tex
\begin{table}[!b]
    \centering
    \small
    \resizebox{\linewidth}{!}{
        \begin{tabular}{c|ccc|c|c}
            \toprule
            & \multicolumn{3}{c|}{\textbf{\textsc{ROUGE-1}}} & \textbf{\textsc{Avg. Number of}} & \textbf{\textsc{Avg. Token}} \\
            & \textbf{P} & \textbf{R} & \textbf{F1} & \textbf{\textsc{Atomic Facts}} & \textbf{\textsc{Length}}\\
            \midrule
            \texttt{Human} & 1.00 & 1.00 & 1.00 & 8.7 & 98.4 \\
            \midrule
            \texttt{Orca-2} & 0.70 & 0.69 & 0.68 & \textbf{8.7} & \textbf{96.3} \\
            \texttt{gpt-3.5-turbo} & \textbf{0.78} & \textbf{0.84} & \textbf{0.79} & 7.8 & 105.0 \\
            \texttt{gpt-3.5-turbo-instruct} & 0.73 & 0.72 & 0.70 & 13.0 & 149.6 \\
            \texttt{Mistral} & 0.63 & 0.62 & 0.61 & 9.6 & 104.1 \\
            \texttt{Zephyr} & 0.51 & 0.60 & 0.52 & 10.1 & 122.0 \\
            \bottomrule
        \end{tabular}
    }
    \caption{Experimental results of generated atomic facts on RoSE dataset. The results with the highest human correlation are highlighted in \textbf{bold}.}
\label{tab:analysis_af}
%\vspace{-5mm}
\end{table}

%% file: FIGURES/fig_ge.tex
\begin{figure}[t]
\centering
\includegraphics[width=\linewidth]{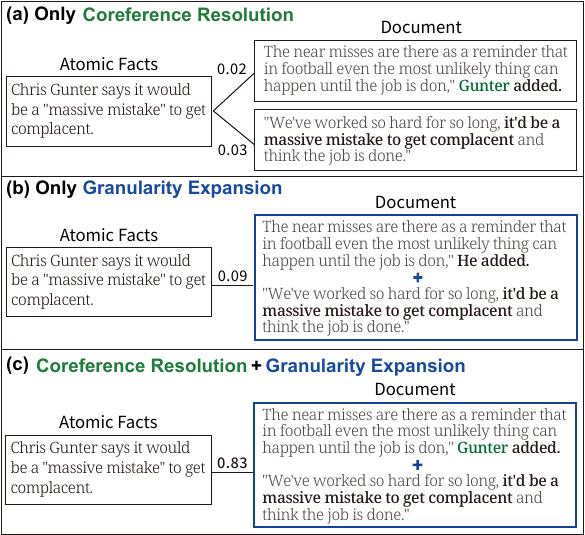}
\caption{The effect of granularity expansions and coreference resolution in real \textsc{AggreFact} dataset. The entailment score of an atomic fact and document sentence with (a) only Coreference Resolution, (b) only Granularity Expansion, and (c) the both.}
\label{fig:GE}
%\vspace{-5mm}
\end{figure}

%% file: TABLES/analysis_granularity.tex
\begin{table}[!t]
    \centering
    \small
    \resizebox{\linewidth}{!}{
        \begin{tabular}{l|cc|c|c}
            \toprule
            \textbf{Doc. Max Granularity} & \begin{tabular}[c]{@{}c@{}}\textbf{\textsc{AggreFact-}}\\\textbf{\textsc{Cnn-FtSota}}\end{tabular} & \begin{tabular}[c]{@{}c@{}}\textbf{\textsc{AggreFact-}}\\\textbf{\textsc{XSum-FtSota}}\end{tabular} & \begin{tabular}[c]{@{}c@{}}\textbf{\textsc{Avg}}\end{tabular} & \begin{tabular}[c]{@{}c@{}}\textbf{s/it}\end{tabular} \\
            \midrule
            One Sent.      & 72.2$\pm$2.8 & 66.3$\pm$1.9 & 69.25 & 2.49 \\
            Two Sent.      & 71.0$\pm$3.2 & 69.3$\pm$2.0 & 70.15 & 2.53 \\
            Three Sent.    & \textbf{72.6}$\pm$3.0 & 69.3$\pm$1.9 & 70.95 & 2.64 \\
            Four Sent.     & 72.1$\pm$3.1 & \textbf{70.0}$\pm$1.8 & \textbf{71.05} & 2.80 \\
            \bottomrule
        \end{tabular}
    }
    \caption{\textbf{Size of granularity choice} in granularity expansion on \textsc{AggreFact-FtSota} split. %We tested four granularity sizes on the document side for the additional lookup of an atomic fact which scored \textit{too low} entailment score. 
    s/it indicates seconds per iteration for the inference of an NLI model.}
\label{tab:analysis_granularity}
%\vspace{-5mm}
\end{table}

%% file: TABLES/analysis_coref.tex
\begin{table}[]
    \centering
    \small
    \resizebox{\linewidth}{!}{
        \begin{tabular}{cc|cc|c}
            \toprule
            \textbf{Atomic Facts} & \textbf{Doc} & \textbf{\textsc{Cnn}} & \textbf{\textsc{XSum}} & \textbf{\textsc{Avg}} \\
            \midrule
            \multirow{2}{*}{Original} & Original   & 63.2$\pm$2.3 & 66.4$\pm$1.8  & 64.8 \\
            & Coref Resolved      & 65.7$\pm$3.4 & \textbf{67.8}$\pm$2.0  & 66.7(\textit{+1.95}) \\
            \midrule
            \multirow{2}{*}{Coref Resolved} & Original  & 66.2$\pm$3.4 & 66.6$\pm$1.9  & 66.4 \\
            & Coref Resolved           & \textbf{72.2}$\pm$2.7 & 66.3$\pm$1.9  & \textbf{69.2}(\textit{+2.85}) \\
            \bottomrule
        \end{tabular}
    }
    \caption{\textbf{Effect of coreference resolution} of document and atomic facts on \textsc{AggreFact-FtSota} splits before the process of granularity expansion.} %Coreference resolved either atomic fact or document both show an increase in the average.}
\label{tab:analysis_coref}
%\vspace{-5mm}
\end{table}

%% file: FIGURES/fig_case.tex
\begin{figure}[t]
\centering
\includegraphics[width=\linewidth]{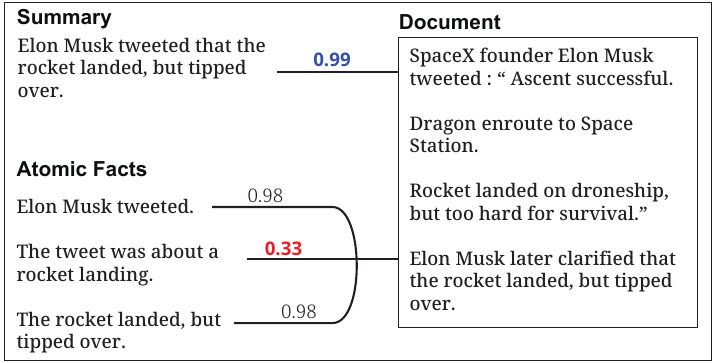}
\caption{Drawbacks of atomic fact level evaluation versus the sentence level evaluation. The numbers represent the maximum NLI entailment scores obtained by comparing each sentence and atomic fact with the source document on a sentence-wise basis.}
\label{fig:case}
\end{figure}

%% file: 5_conclusion.tex
\section{Conclusion}
In this work, we propose a novel method, \text{\metric}, in detecting summary factual inconsistency.
Our approach decomposes summaries into \textit{atomic facts} and conducts a sentence-wise comparison with the document, and achieves state-of-the-art performance on the \textsc{AggreFact} benchmark dataset.
Also, our proposed system has a higher interpretability due to its ability to precisely identify which parts of a summary are factually inaccurate by breaking it down into \textit{atomic facts}.
Furthermore, we analyze the necessity and significance of coreference resolution and granularity expansion in the context of summary factual consistency checking.
%We hope this analysis spurs further research in the field.

%% file: 6_limitations.tex
\section*{Limitations}
Our proposed method is quite time-consuming.
Notably, during the coreference resolution phase, we leverage 11B model.
% Following this, the method involves an additional stage of atomic fact decomposition using an LLM.
% Consequently, this process requires more time than other metrics.
This process requires more time than other factual consistency checking systems.
The practical applicability of \text{\metric} in real-time settings remains to be determined.

Furthermore, our research was limited to summaries based on articles and news domains.
We did not verify the effectiveness of \text{\metric} in other domains such as dialogue summarization~\cite{tang2024tofueval} or medical summarization~\cite{wang-etal-2023-automated}.
Additionally, our study was confined to English-language data.
The validity of \text{\metric} needs to be assessed in datasets based on other languages.

Despite these limitations, we believe our method paves a new path in the area of summarization factual consistency detection.
This work could be a significant contribution to the field, pending further validation across varied domains and languages.

%% file: 7_ethics.tex
\section*{Ethics Statement}
This work uses English document summarization dataset, \textsc{AggreFact}.
This dataset is publicly available online.
We also provided adequate citations for the papers and sources we consulted in writing our paper.
Our work may have implications for society in terms of preventing the spread of inaccurate information, as it deals with factual consistency checking.

% This paper utilizes the English document summary dataset, \textsc{AggreFact}. This dataset is publicly available online.
% We have also cited the papers and sources we used in writing our paper.

%% file: 8_acknowledgement.tex
\section*{Acknowledgement}
This research was supported by the Chung-Ang University Research Grants in 2023. This research was partly supported by Institute for Information \& Communications Technology Planning \& Evaluation (IITP) through the Korea government (MSIT) under Grant No. 2021-0-01341 (Artificial Intelligence Graduate School Program (Chung-Ang University)).

%% file: 9_appendix.tex
\cleardoublepage
\appendix

\section{Prompt for Atomic Facts Decomposition}
\label{sec:prompt_example}
The prompt for atomic fact decomposition in shown in Table~\ref{tab:appx_prompt}.
The examples given in the prompt are similarly used in other LLMs.

\section{Details on Baselines}
\label{sec:exp_details_baselines}
In this section, we present the implementation details of \textsc{FActScore} and \textsc{FacTool}, which have been integrated into our experimental baseline.
For decomposing atomic facts, \textsc{FActScore} uses the \texttt{gpt-3.5-turbo-instruct} model, and the QA process is conducted using \texttt{gpt-3.5-turbo}, with prompts exactly as specified in the paper\footnote{https://github.com/shmsw25/FActScore}.
We gave 1 point for each answer that is answered ture and then divided by the total number of atomic facts:
\begin{equation}
\begin{gathered}
score = \frac{1}{|A|}\sum_{a \in A}\mathbb{I}[\text{~a is True~}] \\
\end{gathered}
\end{equation}

Similar to \textsc{FActScore}, \textsc{FacTool} employs \texttt{gpt-3.5-turbo} for both the \textit{claim} extraction and the QA tasks, again using prompt directly from the paper\footnote{https://github.com/GAIR-NLP/factool}.

\section{Details on the Usage of Large Language Models}
\label{sec:exp_details_llms}
We report on the details and \texttt{Huggingface} links of LLMs used in Section~\ref{sec:experiments}.
We employed Orca-2-7B model\footnote{\url{https://huggingface.co/microsoft/Orca-2-7b}} for experiments in \textsc{AggreFact} benchmark dataset.
For Zephyr, we used Zephyr-7B-beta~\footnote{\url{https://huggingface.co/HuggingFaceH4/zephyr-7b-beta}}, while for Mistral, we used Mistral-7B-instruct-v0.2~\footnote{\url{https://huggingface.co/mistralai/Mistral-7B-Instruct-v0.2}}.
Additionally, we used ChatGPT version of \texttt{gpt-3.5-turbo-0125}.

\section{Details on the Usage of NLI Model}
\label{sec:exp_details_nli}
In this study, we tried to analyze the effect of our proposed atomic fact level decomposition instead of using entire sentences.
To ensure a fair comparison of our approach with \textsc{SummaC}, which demonstrated the best performance using whole sentences, we employed the same NLI model that was utilized in \textsc{SummaC}\footnote{\url{https://huggingface.co/tals/albert-xlarge-vitaminc-mnli}}.
The model has been trained on the conventional NLI datasets SNLI~\cite{bowman-etal-2015-large}, MNLI~\cite{williams-etal-2018-broad}, ANLI~\cite{nie-etal-2020-adversarial}, and also on VitaminC~\cite{schuster-etal-2021-get}.

In Table~\ref{tab:appx_nli}, we present the performance results of various NLI models. 
Specifically, we have included the results for \texttt{DeBERTa-large-mnli}\footnote{\url{https://huggingface.co/MoritzLaurer/DeBERTa-v3-large-mnli-fever-anli-ling-wanli}} and \texttt{RoBERTa-large-pyrxsum}\footnote{\url{https://huggingface.co/shiyue/roberta-large-pyrxsum}}.
The average performance scores for \texttt{DeBERTa} and \texttt{RoBERTa} are 68.7 and 68.5, respectively.
Although these scores are lower than that of \texttt{ALBERT}, they surpass the previous best score of 67.8 achieved by DAE on the FtSota split.

\input{TABLES/appx_nli}

\section{Details on the Usage of Coreference Resolution}
\label{sec:exp_details_cr}
We used MT5-11B model for coreference resolution\footnote{\url{https://huggingface.co/mt5-coref-pytorch/link-append-xxl}}. 
Coreference resolution is the task of identifying all expressions that refer to the same entity within a text.
While recent models perform well on this task, returning a text with resolved coreferences is an entirely different challenge.
We have tested various models, but none have functioned adequately.
A significant issue was the prevalent method of using the first word in a cluster for resolution instead of the entity's name, which frequently resulted in improper replacements with pronouns.
To address this, we slightly modified the code to ensure that where an entity name is available, it replaces pronouns as much as possible\footnote{\url{https://github.com/google-research/google-research/tree/master/coref_mt5}}.
Furthermore, when an adjective or a modifier refers to an entity, we prefixed it with the entity's name followed by a comma.
Table~\ref{tab:appx_cr} illustrates these modifications.
By enhancing coreference resolution in this manner, we were able to capture more comprehensive atomic facts without omitting critical information.

\section{Details on Granularity Expansion}
\label{sec:exp_details_ge}
In Section~\ref{sec:nli_score}, we set the criterion for granularity expansion as $\max(e, c, n) !=e$.
This criterion was chosen because it intuitively signifies a lack of entailment.
Notably, $\max(e, c, n) !=e$ is equivalent to $!(e>c~\&~e>n)$, and thus, we also conducted experiments using the $!(e>c~\|~e>n)$ condition.
Table~\ref{tab:appx_ge_con} presents the results of these experiments.
\input{TABLES/appx_ge_con}

\section{Details on Computing Content Similarity}
\label{sec:exp_details_ccs}
The content similarity (ROUGE-1) in Table~\ref{tab:analysis_af} was conducted using the following equation:

\begin{equation}
    \frac{1}{N_{data}} \sum_{N_{data}} \frac{1}{N_c} \sum_{i=1}^{N_c} \max_{j=1}^{N_g} \left( \text{ROUGE}(c_i, g_j) \right)
\end{equation}
where $c$ denotes LLM generated atomic facts and $g$ denotes human-written atomic facts.

\section{Other Details}
\label{sec:exp_details_other}
In this section, we report the differences observed when splitting text into sentences using NLTK~\cite{bird2009} and Spacy~\cite{honnibal2020}.
We utilized NLTK sentence splitter in \text{\metric}.
The results of the experiments are presented in Table~\ref{tab:appx_sentence}.
\input{TABLES/appx_sentence}

\clearpage
\input{TABLES/appx_prompt}

\clearpage
\input{TABLES/appx_cr}

%% file: TABLES/appx_nli.tex
\begin{table}[h]
    \centering
    \resizebox{\linewidth}{!}{
        \begin{tabular}{l|cc|c}
            \toprule
            \textbf{NLI Model} & \begin{tabular}[c]{@{}c@{}}\textbf{\textsc{AggreFact-}}\\\textbf{\textsc{Cnn-FtSota}}\end{tabular} & \begin{tabular}[c]{@{}c@{}}\textbf{\textsc{AggreFact-}}\\\textbf{\textsc{XSum-FtSota}}\end{tabular} & \begin{tabular}[c]{@{}c@{}}\textbf{\textsc{Avg}}\end{tabular} \\
            \midrule
            ALBERT & \textbf{72.6}$\pm$3.0 & 69.3$\pm$1.9 & \textbf{71.0} \\
            DeBERTa & 67.3$\pm$3.0 & \textbf{70.1}$\pm$1.9 & 68.7 \\
            RoBERTa & 70.5$\pm$3.0 & 66.5$\pm$1.9 & 68.5 \\
            \bottomrule
        \end{tabular}
    }
    \caption{Performance of different NLI models on \textsc{AggreFact-FtSota} split.}
\label{tab:appx_nli}
% \vspace{-5mm}
\end{table}

%% file: TABLES/appx_ge_con.tex
\begin{table}[]
    \centering
    \resizebox{\linewidth}{!}{
        \begin{tabular}{l|cc|c}
            \toprule
            \textbf{Condition} & \begin{tabular}[c]{@{}c@{}}\textbf{\textsc{AggreFact-}}\\\textbf{\textsc{Cnn-FtSota}}\end{tabular} & \begin{tabular}[c]{@{}c@{}}\textbf{\textsc{AggreFact-}}\\\textbf{\textsc{XSum-FtSota}}\end{tabular} & \begin{tabular}[c]{@{}c@{}}\textbf{\textsc{Avg}}\end{tabular} \\
            \midrule
            !(e>c \& e>n) & \textbf{72.6}$\pm$3.0 & \textbf{69.3}$\pm$1.9 & \textbf{71.0} \\
            !(e>c || e>n) & 71.1$\pm$2.9 & 68.7$\pm$1.9 & 69.9 \\
            \bottomrule
        \end{tabular}
    }
    \caption{Granularity Expansion condition choice on \textsc{AggreFact-FtSota} split.}
\label{tab:appx_ge_con}
% \vspace{-5mm}
\end{table}

%% file: TABLES/appx_sentence.tex
\begin{table}[h]
    \centering
    \resizebox{\linewidth}{!}{
        \begin{tabular}{l|cc|c}
            \toprule
            \textbf{Sentence Splitter} & \begin{tabular}[c]{@{}c@{}}\textbf{\textsc{AggreFact-}}\\\textbf{\textsc{Cnn-FtSota}}\end{tabular} & \begin{tabular}[c]{@{}c@{}}\textbf{\textsc{AggreFact-}}\\\textbf{\textsc{XSum-FtSota}}\end{tabular} & \begin{tabular}[c]{@{}c@{}}\textbf{\textsc{Avg}}\end{tabular} \\
            \midrule
            Spacy & 72.5$\pm$3.4 & 67.0$\pm$2.0 & 69.8 \\
            NLTK & \textbf{72.6}$\pm$3.0 & \textbf{69.3}$\pm$1.9 & \textbf{71.0} \\
            \bottomrule
        \end{tabular}
    }
    \caption{Sentence splitter choice on \textsc{AggreFact-FtSota} split.}
\label{tab:appx_sentence}
% \vspace{-5mm}
\end{table}

%% file: TABLES/appx_prompt.tex
\begin{table*}[!t]
{\scriptsize
\centering
\setlength{\tabcolsep}{5pt}
\renewcommand{\arraystretch}{1.2}

% \resizebox{2.05\columnwidth}{!}{
\resizebox{\linewidth}{!}{

\begin{tabular}{p{16cm}}
\toprule
\textbf{Input Prompt} \\
\midrule
% You are Orca, an AI language model created by Microsoft.
You are a helpful assistant. Please give me a list of atomic facts of the following texts. 
\\
\\ lisa courtney, of hertfordshire, has spent most of her life collecting pokemon memorabilia.
\\ - Lisa Courtney is from Hertfordshire.
\\ - Lisa Courtney has spent most of her life collecting Pokémon memorabilia.
\\ 
\\ prince jan zylinski said he was fed up with discrimination against poles living in britain.
\\ - Prince Jan Zylinski made a statement.
\\ - The statement made by Prince Jan Zylinski was about discrimination.
\\ - The statement made by Prince Jan Zylinski was regarding Poles living in Britain.
\\ - Prince Jan Zylinski expressed feeling fed up with this type of discrimination.
\\
\\ no charges were filed, there will be no travel ban.
\\ - No charges were filed.
\\ - There will be no travel ban.
\\
\\ rudd has pleaded guilty to threatening to kill and possession of drugs in a court.
\\ - Rudd has pleaded guilty.
\\ - Rudd has pleaded guilty to threatening to kill.
\\ - Rudd has pleaded guilty to possession of drugs.
\\
\\ Lee made his acting debut in the film The Moon is the Sun's Dream (1992), and continued to appear in small and supporting roles throughout the 1990s.
\\ - Lee made his acting debut in The Moon is the Sun's Dream.
\\ - The Moon is the Sun's Dream is a film.
\\ - The Moon is the Sun's Dream was released in 1992.
\\ - After Lee's acting debut, he appeared in small and supporting roles throughout the 1990s.
\\
\\ In 1963, Collins became one of the third group of astronauts selected by NASA and he served as the back-up Command Module Pilot for the Gemini 7 mission.
\\ - Collins became an astronaut.
\\ - Collins became one of the third group of astronauts selected by NASA in 1963.
\\ - Collins served as the back-up Command Module Pilot for the Gemini 7 mission.
\\
\\ In addition to his acting roles, Bateman has written and directed two short films and is currently in development on his feature debut.
\\ - Bateman has acting roles.
\\ - Bateman has written two short films.
\\ - Bateman has directed two short films.
\\ - Bateman is currently in development on his feature debut.
\\
\\ Michael Collins (born October 31, 1930) is a retired American astronaut and test pilot who was the Command Module Pilot for the Apollo 11 mission in 1969.
\\ - Michael Collins was born on October 31, 1930.
\\ - Michael Collins is retired.
\\ - Michael Collins is an American.
\\ - Michael Collins was an astronaut.
\\ - Michael Collins was a test pilot.
\\ - Michael Collins was the Command Module Pilot for the Apollo 11 mission in 1969.
\\
\\ {\textit{Summary Sentence}}
\\
\bottomrule
\end{tabular}
}
\caption{Prompt used to generate atomic facts from coreference resolved summary in Section~\ref{sec:atomicfact_decomp}. We employed 8-shot learning to enhance the model's performance.}
\label{tab:appx_prompt}
\vspace{-5mm}
}
\end{table*}

% \begin{table}[!t]
% \centering
% \setlength{\tabcolsep}{5pt}
% \renewcommand{\arraystretch}{1}
% \resizebox{\linewidth}{!}{
% \begin{tabular}{p{8cm}}
% \toprule
% \textbf{Input Prompt} \\
% \midrule
% % You are Orca, an AI language model created by Microsoft.
% You are a helpful assistant. Please give me a list of atomic facts of the following texts. 
% \\
% \\ lisa courtney, of hertfordshire, has spent most of her life collecting pokemon memorabilia.
% \\ - Lisa Courtney is from Hertfordshire.
% \\ - Lisa Courtney has spent most of her life collecting Pokémon memorabilia.
% \\ 
% \\ prince jan zylinski said he was fed up with discrimination against poles living in britain.
% \\ - Prince Jan Zylinski made a statement.
% \\ - The statement made by Prince Jan Zylinski was about discrimination.
% \\ - The statement made by Prince Jan Zylinski was regarding Poles living in Britain.
% \\ - Prince Jan Zylinski expressed feeling fed up with this type of discrimination.
% \\
% \\ \textbf{... ( Skip 6 Examples ) ...}
% \\
% \\ {\textit{Summary Sentence}}
% \\
% \bottomrule
% \end{tabular}
% }
% \caption{Prompt used to generate atomic facts from coreference resolved summary in Section~\hyperref[sec:atomicfact_decomp]{3.2}. We employed 8-shot learning to enhance the model's performance.}
% \label{tab:prompt}

% %\vspace{-5mm}
% \end{table}

%% file: TABLES/appx_cr.tex
\begin{table*}[h]
    \centering
    \small
    % \begin{tabularx}{\textwidth}{l|l|X}
    \begin{tabular}{l|l|l}
\toprule
\multicolumn{2}{l|}{\textbf{Original Text}} & \textbf{The 27-year-old} joined spurs from manchester city in 2011.\\
\midrule
% \multirow{6}*{Original Coreference Resolution} 
\multirow{6}*{\textbf{Others}} 
& \textbf{Coref Resolved Text} & {\bf\color{red} Emmanuel Adebayor} joined spurs from manchester city in 2011. \\ 
\cmidrule{2-3}

& Atomic Fact \#1 & Emmanuel Adebayor joined spurs. \\
& Atomic Fact \#2 & Emmanuel Adebayor joined spurs from manchester city. \\
& Atomic Fact \#3 & Emmanuel Adebayor joined spurs in 2011. \\
\midrule

\multirow{6}*{\textbf{Ours}}
& \textbf{Coref Resolved Text} & {\bf\color{blue} Emmanuel Adebayor, the 27-year-old} joined spurs from manchester city in 2011. \\ 
\cmidrule{2-3}

&\textbf{Atomic Fact \#1} & {\bf\color{blue} Emmanuel Adebayor is 27-year-old.} \\
& Atomic Fact \#2 & Emmanuel Adebayor joined spurs. \\
& Atomic Fact \#3 & Emmanuel Adebayor joined spurs from manchester city. \\
& Atomic Fact \#4 & Emmanuel Adebayor joined spurs in 2011. \\
\bottomrule
    \end{tabular}
    % \end{tabularx}
    % \vspace{-0.2cm}
    \caption{Our distinct approach for coreference resolution. The original text is coreference resolved by two ways, which are \textbf{Others} and \textbf{Ours}. We ensure that critical information is preserved while generating atomic facts by prefixing modifiers with the names of entities during the coreference resolution.}
    \label{tab:appx_cr}
    % \vspace{-0.2cm}
\end{table*}